\documentclass{article}
\usepackage{float}
\usepackage[preprint]{neurips_2020}
\usepackage{enumitem}
\usepackage{geometry}
\usepackage{graphicx}
\usepackage{minitoc}
\usepackage{longtable}
\usepackage{qtree}
\usepackage{tcolorbox}
\usepackage{tikz}
\usepackage{tabularx}
\usepackage{booktabs}
\usepackage{siunitx}
\usepackage{ragged2e}
\usepackage{rotating}
\usepackage{adjustbox}
\usepackage{caption}
\usepackage[T1]{fontenc}
\usepackage{pdflscape}
\usepackage{nameref}
\usepackage{pdfpages}
\usepackage[multiple]{footmisc}
\usepackage{float}
\newcolumntype{R}[1]{>{\RaggedRight}p{#1}}
\usetikzlibrary{backgrounds} 
\usepackage{amssymb}
\usetikzlibrary{arrows.meta, positioning, calc}
\usepackage[x11names,dvipsnames]{xcolor}
\usepackage{verbatim}
\usepackage{subcaption}
\usepackage{array}
\usetikzlibrary{positioning, arrows.meta, shapes.geometric, fit, shadows}
\usepackage{natbib}
\usepackage{url}
\usepackage[hidelinks]{hyperref}
\usepackage{doi}
\usepackage{mathtools}
\usepackage{xspace}
\usepackage{cleveref}
\usepackage{svg}
\usepackage{transparent}
\usepackage[toc,title,page]{appendix}
\definecolor{orangefull}{RGB}{230, 159, 0}
\newcommand*{\eg}{e.g.\@\xspace}
\newcommand*{\ie}{i.e.\@\xspace}

\usepackage{siunitx}
\usepackage[pagewise]{lineno}  
\colorlet{orange}{orangefull!20!white}
\definecolor{bluefull}{RGB}{86, 180, 233}
\colorlet{blue}{bluefull!20!white}
\definecolor{green}{RGB}{0, 158, 115}
\definecolor{checkblue}{RGB}{86,180,233}

\colorlet{green}{green!20!white}
\title{How Can Machine Learning Emulators Best Support Climate Science?}

\begin{document}
\definecolor{checkdark}{RGB}{60,60,60}
\author{%
  Luca Schmidt \thanks{Equal contribution, other authors in alphabetical order}\\
  Cluster of Excellence Machine Learning\\
  University of Tübingen\\
  \texttt{luca.schmidt@uni-tuebingen.de} \\
  \And
  Nina Effenberger \footnotemark[1] \\
  Institute for Atmospheric and Climate Science \\
    ETH Zurich \\
    \texttt{nina.effenberger@env.ethz.ch} \\
    \And 
   Vitus Benson \\
  Max Planck Institute for Biogeochemistry \\
  \texttt{vbenson@bgc-jena.mpg.de}
 \And
   Philine L. Bommer\\
   School of Geosciences\\
     University of Edinburgh \\
    \texttt{philine.bommer@ed.ac.uk} \\
   \And
   Robert Brunstein\\
   Otto-von-Guericke-Universität Magdeburg\\
     \texttt{robert.brunstein@ovgu.de} \\
   \And 
   Mikel N. Legasa \\
  Laboratoire des Sciences du Climat et de l’Environnement (LSCE) \\ CEA/CNRS/UVSQ, Université Paris Saclay \\
  Institut Pierre-Simon Laplace (IPSL) \\
  \texttt{mikel.legasa@lsce.ipsl.fr}
   \And
   Maxim Samarin \\
   Swiss Data Science Center \\
   EPFL and ETH Zurich \\
   \texttt{maxim.samarin@sdsc.ethz.ch} \\
   \And
   Maybritt Schillinger \\
   Seminar for Statistics \\
   ETH Zurich \\
   \texttt{maybritt.schillinger@stat.math.ethz.ch} \\
}

\maketitle

\begin{abstract}
    For decades, physics-based climate models have been used to provide insights for climate decision-making. Their application is, however, constrained by significant computational and technical demands. Machine learning (ML) emulators offer a way to reduce these high computational costs; yet, it remains challenging to use ML emulators effectively in climate research. In practice, climate scientists often bypass emulators altogether, and machine learning researchers frequently develop them as methodological showcases without proving their practical utility. The reasons are diverse, ranging from limited accessibility and a lack of specialized knowledge to broader concerns about the physical grounding of ML methods. Here, we discuss limitations and introduce a framework for guiding emulator development, considering both climate science and machine learning perspectives. We argue that designing easy-to-adopt emulators that address clearly defined tasks and demonstrate their reliability is essential. This offers a promising path towards making machine-learning approaches more relevant and usable for applied climate research.
\end{abstract}

\section{Introduction}
Climate models represent our most advanced understanding of the climate system. Their insights into current and future climate change guide mitigation, adaptation, and other societal and economic decisions \citep{masson2021climate}. Climate model simulations, however, are computationally demanding and technically complex, requiring substantial computational resources and specialized expertise \citep{balaji2017cpmip}. 
The increasing demand for climate information, along with the cost and difficulty of running physical climate models, calls for more efficient modeling approaches. Machine learning-based climate model emulators fill this gap. They can approximate specific model parameterizations or 
act as full statistical and machine learning surrogates (\eg \citep{leach2021fairv2, bouabid2026score, hickman2025causal}). All such models share the goal of approximating key elements of advanced physical climate models.

In recent years, interest in climate model emulation has increased in both climate science and machine learning, accompanied by rapid methodological development--especially within the methodological community \citep[\eg][]{fowler2025downscaling, rampal_artificial_2024}.
For the climate science community, emulators provide an efficient way to augment simulations \citep{kendon2025potential}, quantify and reduce uncertainty \citep{watson2021machine}, and generate climate information that would be infeasible with full climate models. For the machine learning community, particularly in computer vision, climate emulation offers a compelling scientific challenge and an opportunity to demonstrate the value of advanced modeling techniques in socially relevant applications \citep{watson2021machine}.

Despite this shared motivation, the two communities remain surprisingly disconnected. Their differing perspectives toward emulation are reflected in their workflows: the machine learning community typically develops data-driven models optimized for predefined predictive tasks, whereas climate scientists analyze physics-based simulations to address specific scientific or application-driven questions. These disparate approaches mean that the paths of methodological developers and applied users rarely intersect \textit{by chance}. Research inertia may further contribute to this divide. For example, climate researchers tend to rely on established \emph{gold standard} datasets like ERA5 \citep{hersbach2020era5} or CMIP6 \citep{eyring2016overview, adams2018sampling}, while other approaches are often considered only once traditional models reach their limits. More generally, they may prefer working with familiar datasets and models whose limitations are well understood, rather than engaging with the \emph{unknown unknowns} introduced by unfamiliar approaches.\footnote{This is a well-researched topic in social science, where it has been found that people prefer a known risk over ambiguity \citep{ellsberg1961risk}.}
As a result, rapid progress in machine learning climate model emulators has not resulted in comparable uptake by users \citep{fowler2007linking, tebaldi2007use, fowler2025downscaling}. Many emulators, therefore, remain unused in decision-making contexts; not because they fall short, but because developers and users lack a shared framework for communication, evaluation, and practical guidance. This calls into question the added value of much of this modeling effort \citep{fowler2025downscaling}.

Closer collaboration could benefit both sides. For the climate science community, the advantages may seem particularly clear: machine learning emulators can substantially reduce computational costs \citep{doury2023regional} and even enable capabilities that are infeasible with physical models. Recent efforts, such as AIMIP Phase 1, are a promising first step in this direction \citep{henn2026aimip}.
The machine learning community likewise has an interest in seeing its models used in real-world applications. Ultimately, increasing societal relevance gives these models a purpose.
This view aligns with the philosophical idea of \textit{adequacy-for-purpose} \citep{parker2020model}, in which a model derives its meaning only from its suitability for a particular task and context, while abstract accuracy metrics alone are insufficient to judge its value.
Closer interaction between emulator developers and users could also enable the creation of a feedback loop between emulators and physical simulations. For example, by enabling large ensembles and sensitivity analyses, machine-learning models can identify regions where additional
physical simulations would most effectively reduce uncertainty \citep{van2026rewiring}. These insights could guide more targeted physical simulation efforts, which in turn provide improved training data for emulators. 

In this paper, we address the disconnect between the technical and applied communities. We first clarify how the term \textit{climate model emulator} is understood in each field and compare the perspectives that shape how climate scientists and machine learning researchers approach emulation. In \Cref{sec:development}, we establish a systematic framework for developing scientifically motivated, trustworthy, and accessible machine-learning emulators. Ultimately, we aim to support more targeted emulator development and simplify their uptake in decision-making contexts.

\section{Definition: What are emulators and what can they be used for?}
\label{sec:definition}

\subsection{What are climate model emulators?}
The IPCC defines emulation as ``reproducing the behavior of complex, process-based models (namely, Earth System Models (ESMs)) via simpler approaches'' \citep{matthews2021annex}. Models developed for this purpose are often referred to as simple climate models because they represent the Earth system using simplified physical process formulations. Simple climate models typically aim to reproduce key decision-relevant statistics of climate variables rather than the full dynamical evolution of the climate system \citep{leach2021fairv2, romero2025review}.

\begin{figure}[h]
    \centering
\begin{tikzpicture}[
    node distance=0.5cm, 
    container/.style={draw, minimum width=3.2cm, minimum height=3.6cm, align=center, thick, fill=gray!2, rounded corners=3pt},
    photo/.style={draw, fill=white, inner sep=1pt, thick},
    arrow/.style={-{Stealth[scale=0.85]}, line width=1.5pt, color=black}
]
    \node[container, draw=teal!60, fill=teal!5] (A) at (0,0) {};
    \node[anchor=north, yshift=-5pt] at (A.north) {\textbf{Input}};
    \begin{scope}[shift={(A.center)}, xshift=-0.4cm, yshift=-0.5cm]
        \node[photo] at (0.6, 0.6) {\includegraphics[width=1.3cm, height=0.9cm]{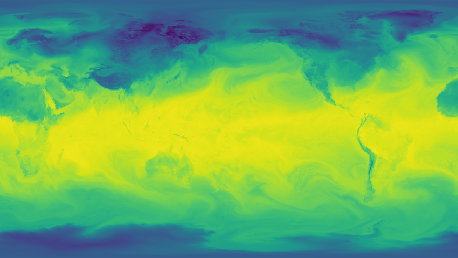}};
        \node[photo] at (0.4, 0.4) {\includegraphics[width=1.3cm, height=0.9cm]{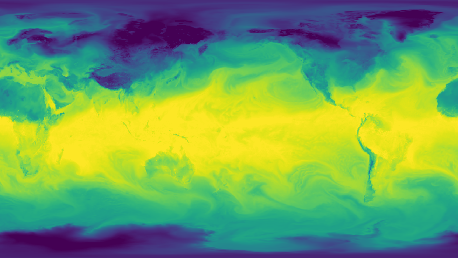}};
        \node[photo] at (0.2, 0.2) {\includegraphics[width=1.3cm, height=0.9cm]{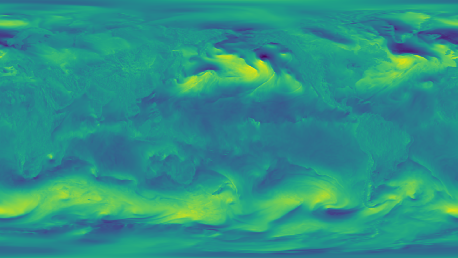}};
        \node[photo] (top) at (0, 0) {\includegraphics[width=1.3cm, height=0.9cm]{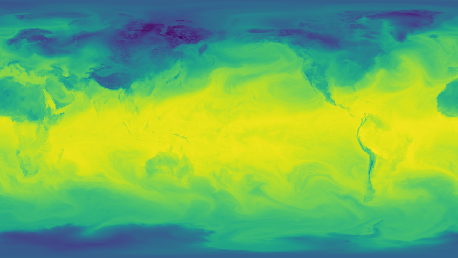}};
    \end{scope}

    \node[inner sep=0pt, align=center, minimum width=1.8cm] (icon) [right=of A] {
        \vspace*{0.4cm} 
        \includegraphics[width=1.4cm]{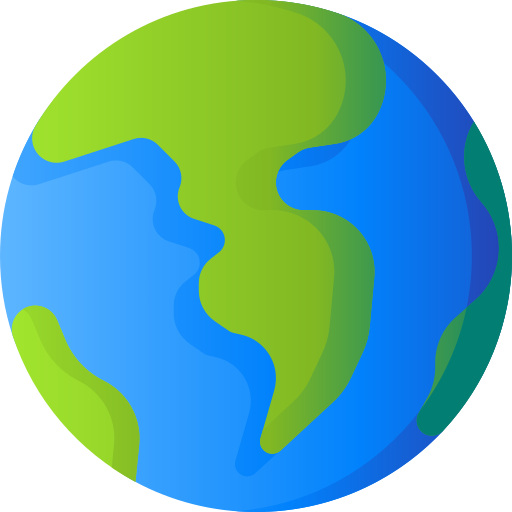} \\[0.1cm]
        \textbf{CM}
    };

    \node[container, draw=teal!60, fill=teal!5, right=of icon] (B) {};
    \node[anchor=north, yshift=-5pt] at (B.north) {\textbf{CM output}};
    \begin{scope}[shift={(B.center)}, xshift=-0.4cm, yshift=-0.5cm]
        \node[photo] at (0.6, 0.6) {\includegraphics[width=1.3cm, height=0.9cm]{fig/t2m_3.png}};
        \node[photo] at (0.4, 0.4) {\includegraphics[width=1.3cm, height=0.9cm]{fig/t2m_max.png}};
        \node[photo] at (0.2, 0.2) {\includegraphics[width=1.3cm, height=0.9cm]{fig/u10_mean.png}};
        \node[photo] (topB) at (0, 0) {\includegraphics[width=1.3cm, height=0.9cm]{fig/t2m_shifted.png}};
    \end{scope}

    \node[container, draw=teal!60, fill=teal!5, right=of B] (C) {
        \textbf{Post-processing} 
    };

    \draw[arrow] (A.east) -- (icon.west);
    \draw[arrow] (icon.east) -- (B.west);
    \draw[arrow] (B.east) -- (C.west);
    
    \draw[arrow, dash pattern=on 10pt off 6pt] ([yshift=0.3cm]B.150) -- node[above, font=\footnotesize\sffamily, text=black, yshift=4pt] {Feedback} ([yshift=0.3cm]A.30);

    \node[draw=teal!60, thick, fill=teal!5, rounded corners=3pt,
          below=1.8cm of icon, 
          xshift=1.85cm, 
          minimum width=11.8cm,
          minimum height=1.3cm,
          inner sep=7pt,
          align=center] (bottombox) {
              \textbf{Emulator}\\
            \includegraphics[width=1.3cm]{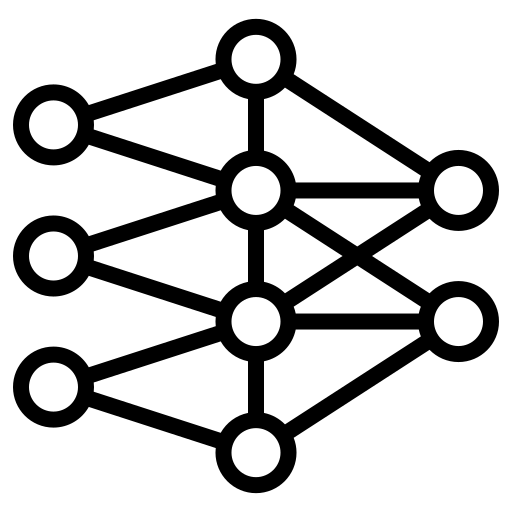}
    };

    \draw[arrow] (A.south) -- (A.south |- bottombox.north);
    \draw[arrow] (bottombox.north -| B.south) -- node[right, font=\footnotesize\sffamily, text=black, xshift=2pt] {replication} (B.south);
    \draw[arrow] (bottombox.north -| C.south) -- node[right, font=\footnotesize\sffamily, text=black, xshift=2pt] {extension} (C.south);

\begin{pgfonlayer}{background}
    \node[draw=gray!40, fill=gray!7, rounded corners=6pt, inner sep=6pt,
          fit=(A)(B)(C)(icon)(bottombox)] {};
\end{pgfonlayer}
\end{tikzpicture}
   \caption{Climate models (\textit{CMs}) use physical equations to transform input data (top left) into consistent outputs (center), which can recursively serve as inputs for subsequent modeling steps (\textit{Feedback}). Data-driven emulators can replace these physical components (\textit{replication}), typically trained on paired physical model simulations. Beyond standard replication, emulators can also extend the modeling pipeline by directly integrating post-processing, such as bias correction or extreme event simulation (\textit{extension}).}
    \label{fig:emulators}
\end{figure}

In contrast, our main focus here is on machine learning models that \textit{emulate} physical climate models by learning statistical relationships directly from data (see \Cref{fig:emulators}), which we refer to as ML emulators or simply emulators. These emulators may be trained on global or regional climate model output, or exclusively on observational or reanalysis data, \citep[\eg][]{aich2026conditional}. Examples include radiative forcing exploration \citep[\eg][]{schongart2025high}, climate model parameter tuning \citep[\eg][]{bonnet2025tuning}, and regional climate data generation \citep[\eg][]{schillinger2025enscale, tomasi_can_2025, addison_machine_2026}. All emulators we consider share the goal of representing climate processes at different spatial and temporal scales.

\subsection{What can emulators be used for?}
Depending on their design and training objectives, ML climate model emulators can either replicate the behavior of existing climate models or extend their capabilities. In the most basic case, they act as surrogates for physical models, reproducing their behavior at a lower computational cost. Such surrogate emulators can be validated with the same diagnostics as the original climate model. As they reproduce the system’s behavior, these models will also reproduce biases and face the same validation challenges \citep[\eg][]{rummukainen2016added, fowler2025downscaling,watt2025ace2,guan2024lucie,cresswell2025deep}. ML surrogate emulators can be valuable by making climate information more accessible. For example, their ability to generate climate data on demand can reduce the need to store, download, and process large datasets such as CMIP6, whose estimated volume exceeds $16 \mathrm{PB}$ \citep{durack2025coupled}. 

Other emulators go beyond reproducing existing model output. 
For example, they can combine multiple data sources (climate models or observational datasets) \citep{wan2024regional}, sample rare extreme events more efficiently \citep{kendon2025potential}, perform bias correction, enhance temporal or spatial resolutions \citep{harder2026benchmarking}, quantify uncertainties through probabilistic approaches \citep{schillinger2025enscale, rampal2025downscaling}, project climate outcomes for new emission scenarios \citep{clyne2026emulating}, and learn complex relationships that are difficult to parameterize within traditional physical frameworks \citep{rasp2018deep}. However, such extensions also make validation and interpretation more challenging because the emulator’s outputs no longer strictly align with the outputs of a physical climate model.\footnote{The boundary between surrogates and emulators with extended functionality is blurred and complicates their classification alongside traditional physical models.}

ML-based emulators can also complement complex traditional climate models \citep{balaji2017cpmip}. Such frameworks create opportunities for hybrid approaches, such as NeuralGCM \citep{kochkov2024neural}, where learned components are embedded within physically grounded modeling frameworks. 
This allows climate model emulators to incorporate physical knowledge \citep{reichstein2019deep} and may also create opportunities for scientific discovery \citep{ qian2022d, beucler2025distilling}.

\definecolor{one}{RGB}{230, 159, 0}
\definecolor{two}{RGB}{86, 180, 233}
\definecolor{three}{RGB}{0, 158, 115}
\definecolor{four}{RGB}{0, 114, 178}

\section{Two perspectives on emulators}
A key challenge is that machine learning and climate science follow different research workflows \citep{wu2025bridging}. Machine learning research (top row in \Cref{fig:combinedworkflow}) is typically predictive task-driven: the goal is to develop and validate a model optimized for a predefined task, such as refining spatial data resolution. Often, ML research generates scientific value through technical innovations even before a specific domain application is addressed. For instance, the Transformer architecture \citep{vaswani2017attention} originated as a technical innovation and has since become a key tool in ML weather forecasting \citep{bi2023accurate, nguyen2023climax}. 
By contrast, the key focus in climate research is understanding physical processes, quantifying uncertainty, and answering broader scientific or societal questions \citep{wu2025bridging}. 
Studies usually begin with such a question (\eg, how regional wind resources may be affected by climate change), and defining the precise task and selecting appropriate data are themselves part of the research process (bottom row in \Cref{fig:combinedworkflow}).

These different perspectives, the 'what' of the machine learner and the 'why/how' of the climate scientist,  shape how each community approaches emulation. \Cref{fig:combinedworkflow} illustrates how the two research paths\footnote{The binary distinction used here is a deliberate simplification. In practice, many researchers work at the intersection of both fields (including the authors of this paper), and their degree of engagement with either community varies depending on their research environment and context.} relate to each other and highlights potential points of interaction (dashed arrows) where collaboration could facilitate the successful integration of ML emulators into applied climate research. 
\begin{figure}[h]
    \centering
\usetikzlibrary{backgrounds}
\begin{tikzpicture}[
  node distance=1cm and 1cm,
  box/.style={draw=gray!70, rectangle, rounded corners, minimum width=2cm, minimum height=1cm, align=center, fill=gray!8},
  tag/.style={align=center, font=\bfseries},
  >=Stealth
]
  \node[tag] (ml) {ML};
  \node[box] (b1) [right=of ml] {define predictive task};
  \node[box] (b2) [right=of b1] {develop model};
  \node[box] (b3) [right=of b2] {generate data};
  \node[box] (b4) [right=of b3] {validate data};
  \draw[->, thick, color=gray!90] (b1) -- (b2);
  \draw[->, thick, color=gray!90] (b2) -- (b3);
  \draw[->, thick, color=gray!90] (b3) -- (b4);
  \node[tag] (appl) [below=1.5cm of $(b1)!0.5!(b4)$, xshift=-7.9cm] {Climate};
  \node[box] (c1) [right=of appl, xshift=-0.2cm] {develop applied\\research question};
  \node[box] (c2) [right=of c1] {define data\\requirements};
  \node[box] (c3) [right=of c2] {analyze data};
  \node[box] (c4) [right=of c3] {answer research\\question};
  \draw[<->, thick, color=gray!90] (c1) -- (c2);
  \draw[->, thick, color=gray!90] (c2) -- (c3);
  \draw[->, thick, color=gray!90] (c3) -- (c4);
  \draw[<-, thick, dashed, color=gray!90] (b1.south) -- (c2);
  \draw[->, thick, dashed, color=gray!90] (b4.south) -- (c3);

\end{tikzpicture}
\caption{Combined emulator workflow. The top row shows the typical path of a research project in ML, the bottom row shows applied research. The goal of emulator ML research is to develop and validate an emulator to generate data. Research in climate science typically centers around analyzing and validating such data. For this purpose, emulated data can be used but they are usually not necessary.}
    \label{fig:combinedworkflow}
\end{figure}
In most cases, the workflows of ML scientists and applied researchers proceed largely independently and rarely intersect. To better understand this disconnect, we first describe the typical perspectives and workflows in both communities in \Cref{sec:MLview} and \Cref{sec:CSview} before discussing how they can be more effectively connected in \Cref{sec:development}.

\subsection{A machine learning researcher's view}
\label{sec:MLview}
Climate emulation presents an appealing scientific challenge--one where ML can shine by replacing expensive climate model runs with fast surrogate models. 

Machine learning research typically begins with a clearly defined prediction task -- for example, learning a mapping from coarse- to high-resolution climate data -- followed by selecting suitable datasets. 
Yet, considerable effort is then devoted to making the data \textit{ML ready}, \ie, bringing it into the right format. In climate emulation, this naturally makes gridded products like ERA5 reanalysis \citep{hersbach2020era5} a preferred choice: they are readily accessible, standardized, and often treated as proxies for observational reference data. Further, gridded data integrates smoothly into ML pipelines, specifically deep learning based emulators, that typically employ convolutional operators \citep[\eg][]{doury2023regional} that naturally work with gridded data.
 
Once the data is selected and pre-processed, the researcher's attention shifts to choosing a suitable algorithm, training a model, and tuning hyperparameters. Benchmarking plays a central role in this phase, typically relying on standardized, often aggregated performance metrics with the goal of developing a model that ranks favorably against others. The implicit assumption is that the top-performing model will also be the most suitable starting point for a certain downstream application \citep{hardt2025emerging}.

Recent advances in generative models have made them particularly attractive for climate emulation, as they can provide probabilistic predictions, see for example \citep{brenowitz2025climate,singh2026evaluating, addison_machine_2026, legasa_regional_2026}. In practice, they are able to characterize their own uncertainty through the spread of the model’s predictions. However, other sources of uncertainty--such as those related to data selection, predictor choice, model transferability, or non-stationarity--are typically considered separately or remain challenging to quantify in a systematic way \citep{fowler2025downscaling, tebaldi2025emulators}. 


\subsection{A climate scientist's view}
\label{sec:CSview}
In contrast to ML research, applied science is usually motivated by a broader research question; for example, how a weather variable in a certain region is affected by climate change. 

Data collection and pre-processing techniques are informed by this question, drawing on sources such as observational datasets, reanalysis products, and climate model outputs. The researcher usually first defines clear criteria, such as independence, performance, and spread \citep{merrifield2023climate}, and then identifies datasets that best meet them. In practice, no single dataset will be perfect: one may perform best under one criterion while another excels under different criteria. This is particularly challenging in climate science, where relevant time horizons extend far beyond what can be directly verified, and researchers must rely on model simulations that differ in many aspects. Uncertainty quantification is therefore not optional but central. In applied research, data selection is essential; yet, in robust cases, conclusions must remain valid regardless of the datasets and methods used.  

Following the initial data selection, the researcher begins analyzing the data to address the research question. During this phase, exploratory analyses may reveal unexpected patterns or insights that can lead to further investigation, additional data collection, or refinement of the original data selection or research question. This iterative cycle of analysis and validation continues until the researcher is satisfied that the data and findings are robust. Finally, the results are interpreted, visualized, and communicated, ensuring that the findings are both clear and reproducible. 

But where and how do machine learning emulators fit into this pipeline?

\section{Guiding emulator development} 
\label{sec:development}
Methodological and applied research pathways are often disconnected (\Cref{fig:combinedworkflow}); climate scientists often bypass emulators altogether, and machine learning researchers frequently develop them as methodological advances without proving practical utility.\footnote{\Cref{sec:ace2_references} illustrates this gap using ACE2 \citep{watt2025ace2} as an example. We categorized the 60 papers citing the original ACE2 publication as of 15 June 2026 by citation type and found that only two used ACE2 in a concrete practical application, \ie generate new knowledge beyond existing data.}
Consequently, their performance beyond well-defined benchmark settings remains uncertain, and design choices are frequently guided by relative model rankings rather than a clear assessment of how well an emulator serves a specific application. In addition, modeling assumptions, limitations, and uncertainties are often not communicated in ways that are familiar to applied researchers. 

In the following, we outline the steps needed to connect ML emulators to applied research. To be effective, emulators must do more than perform well from a technical standpoint; they must provide genuine added value to the applied research question. We propose three criteria for a useful emulator: it must address a scientifically-motivated task (adequacy for purpose), perform that task reliably (trustworthiness), and be easy for the broader community to adopt (accessibility), see \Cref{fig:overview-pillars} for an overview. The remainder of this section is structured around the three pillars. We summarize our findings in a checklist that can guide practical implementation in \Cref{fig:checklist}.

\begin{figure}[h]
    \centering
    \newcommand{\sbull}{\raisebox{0.15ex}{\scalebox{0.55}{$\bullet$}}\,}
\begin{tikzpicture}[
    node distance = 0.4cm,
    box/.style = {rectangle, rounded corners, minimum width=5.0cm, minimum height=1cm, align=left, draw=black!70, fill=white, font=\normalsize, drop shadow},
    title/.style = {font=\bfseries\large, color=white, align=center},
    arrow/.style = {-{Stealth[scale=1.2]}, line width=1.5pt, color=gray!80}
]
\node[fill=gray!90, minimum width=5cm, minimum height=0.8cm, rounded corners=2pt] (title1) {\textcolor{white}{\textbf{Adequacy for Purpose}}};
\node[box, below=0.5cm of title1] (prob) {\textbf{Scientifically motivated task}\\ \tiny \sbull Downstream application \& research context \\ \tiny \sbull Identify application-relevant aspects \\ \tiny \sbull Consult applied researcher};
\node[box, below=of prob] (data) {\textbf{Data selection}\\ \tiny \sbull Problem-aligned data (scales, regimes etc.) \\ \tiny \sbull Physically relevant predictors \\ \tiny \sbull Biases and preprocessing documented};
\node[box, below=of data] (eval) {\textbf{Purpose-driven evaluation}\\ \tiny \sbull Application-relevant metrics\\ \tiny \sbull Extremes \& object-based metrics \\ \tiny \sbull Downstream impact tests};
\node[fill=gray!90, minimum width=5cm, minimum height=0.8cm, right=0.6cm of title1, rounded corners=2pt] (title2) {\textcolor{white}{\textbf{Trustworthiness}}};
\node[box, below=0.5cm of title2] (value) {\textbf{Demonstrate added value}\\ \tiny \sbull With context-dependent metrics  \\ \tiny \sbull Compared to meaningful reference};
\node[box, below=of value] (uncert) {\textbf{Understand uncertainties}\\ \tiny \sbull Source of uncertainty \\ \tiny \sbull Interpretation of uncertainty};
\node[box, below=of uncert] (gen) {\textbf{ Generalizability}\\ \tiny \sbull Performance under climate change\\ \tiny \sbull Other aspects (\eg space, other climate models)};
\node[fill=gray!90, minimum width=5cm, minimum height=0.8cm, right=0.6cm of title2, rounded corners=2pt] (title3) {\textcolor{white}{\textbf{Accessibility}}};
\node[box, below=0.5cm of title3] (standard) {\textbf{Open access}\\ \tiny \sbull Provide open access to full workflow \\ \tiny \sbull Use interoperable formats};
\node[box, below=of standard] (docs) {\textbf{Clear documentation}\\ \tiny \sbull Intended use \\ \tiny \sbull Scope and known limitations \\ \tiny \sbull Clear communication to non-ML users};
\node[box, below=of docs] (collab) {\textbf{Be open to collaboration}  \\ \tiny \sbull Leverage frameworks (like Anemoi) \\ \tiny \sbull Show interest in inter-disciplinary collaboration};
\coordinate (bottom) at (eval.south);
\begin{scope}[on background layer]
    \node[fill=gray!10, draw=gray!20, fit=(title1) (eval) (eval.south |- bottom), inner sep=5pt, rounded corners] (bg1) {};
    \node[fill=gray!10, draw=gray!20, fit=(title2) (gen) (gen.south |- bottom), inner sep=5pt, rounded corners] (bg2) {};
    \node[fill=gray!10, draw=gray!20, fit=(title3) (collab) (collab.south |- bottom), inner sep=5pt, rounded corners] (bg3) {};
\end{scope}
\end{tikzpicture}
\caption{Three criteria for guiding the development of useful climate model emulators. A climate model emulator is most likely to support applied research when it addresses a scientifically motivated purpose, is trustworthy within its intended scope, and is accessible to the applied community.}
    \label{fig:overview-pillars}
\end{figure}

\subsection{Adequacy for purpose} 
Because ML climate model emulators vary greatly in their methods, complexity, requirements, and outputs \citep{tebaldi2025emulators}, it can be difficult to assess their suitability for a particular applied task.

The first step in developing an emulator should be to clearly define the specific problem that the emulator aims to solve. This includes identifying relevant predictors, the spatial and temporal domains, and selecting evaluation criteria. While the technical definition of the task lies with the ML developer, it should be informed by the specific needs of the applied researchers who will ultimately use it, as illustrated in \Cref{fig:combinedworkflow}.  
The same applies to data selection, which is frequently shaped by availability and convenience rather than by demonstrated adequacy for a specific application. This tendency is also reflected in the common practice of benchmarking \citep[e.g.,][]{harder2026benchmarking, rampal_cordexbench_2026}, which is valuable for standardized comparisons across models using consistent datasets and metrics \citep{thiyagalingam2022scientific}. At the same time, strong performance in these settings does not guarantee real-world applicability \citep{lutjens2025impact}, and incremental gains in model metrics may not translate into actual advances in understanding climate impact. 
Benchmarking primarily supports the comparison of model architectures under the assumption that data are exchangeable. However, in climate science, data is tied to its physical context, and model performance can strongly depend on the specific site, season, and other factors \citep{fowler2025downscaling}. For example, emulators producing output on daily timescales may be insufficient for applications such as wind power forecasting \citep{effenberger2024mind}.
Therefore, the input data used to train the emulator should undergo systematic checks \citep{kannan2014uncertainty, gutmann2014intercomparison}, with predictor variables chosen to reflect relevant physical processes. 

Model evaluation should go beyond simply comparing average statistics of relevant variables and instead consider the entire emulator workflow \citep{fowler2025downscaling}.
 Evaluations must also account for large internal climate variability \citep{deser2012communication} and should separate natural variability from the forced climate signal, for example, by using sufficiently large ensembles \citep{lutjens2025impact}. Therefore, evaluation metrics should be chosen with the intended application in mind and justified accordingly \citep{kendon2025potential}. Further, evaluating climate model data on a point-by-point basis is of limited value; see also \citet{deser2012communication}. More informative evaluations could examine whether emulators reproduce distributional properties such as variability and extremes \citep{sun2020evaluating}, preserve physically meaningful inter-variable relationships, capture the structure and evolution of coherent weather phenomena through object-based metrics \citep{doury2023regional}, and, for generative models, represent plausible uncertainty ranges for selected scientifically relevant events, e.g., extremes \citep{legasa_regional_2026}.
 Such an evaluation need not be exhaustive across all possible metrics, but it should clearly validate the emulator’s performance relative to its stated scientific purpose. 
 Where emulators are intended to support downstream applications, validation can also be extended to impact assessments; for example, by testing whether emulator-based impacts are consistent with those derived from the original physical models \citep{bouabid2026score}. Finally, comparing results against multiple observational products whenever possible strengthens the robustness of findings \citep{gonzalez2019harmonized, kochkov2024neural}.

\subsection{Trustworthiness}
Another barrier to the uptake of emulators is the lack of trust within the climate science community. This hesitation arises because an emulator's underlying assumptions, sources of uncertainty, and limits of robustness are often poorly understood. Consequently, the added value of these tools, and with it their status relative to traditional physical models, remains ambiguous. Without a consistent framework to quantify their value, it is unclear whether an emulator is trustworthy enough for decision-making.
Added value is not a universal metric; its definition varies significantly across the literature \citep{ fowler2007linking, di2015challenges, legasa_regional_2026}. While it is often quantitatively defined as an emulator's proximity to a reference dataset, true added value is inherently context-dependent. For example, a common pitfall when evaluating downscaling emulators\footnote{Downscaling refers to increasing the resolution of climate data.} is equating the presence of finer-scale spatial structure with added value. However, this finer structure only results in genuine added value if the physically relevant aspects of the simulation are improved \citep{morelli2025climate, di2015challenges}. Consequently, added value is highly sensitive to the chosen metrics, variables, scales, and specific user requirements. 
A good understanding of what constitutes added value can be a key indicator of whether an emulator will be deployed in practice.

Furthermore, understanding the uncertainty across the modeling pipeline is essential for building trust in ML emulators \citep{haynes2023creating}. 
In climate modeling, uncertainty arises from many sources, including model structure and parameterizations, internal variability, emission scenarios, and inter-variable relationships \citep{tebaldi2025emulators, watson2021machine, kendon2025potential, lutjens2025impact}. 
Uncertainty in different observational or reanalysis products can also be substantial and can even exceed that of global climate models \citep{kannan2014uncertainty}. Recent work further shows that the observational reference used for bias adjustment can contribute more uncertainty than the choice of downscaling or bias-correction method itself and may lead to different conclusions for climate change adaptation decisions \citep{lavoie2025importance}.
Emulator design directly shapes which uncertainties are represented and how the resulting outputs must be interpreted. To account for this, uncertainties arising from data choice, preprocessing, or modeling assumptions should be explicitly documented and discussed \citep{fowler2025downscaling,tebaldi2025emulators}. 

Another key question regarding the usefulness of an emulator is whether it remains valid beyond the training and testing conditions. Yet many recent ML studies do not address such robustness under out-of-distribution conditions \citep[\eg][]{mooers2021assessing, yu2025climsim}. In climate applications, this assumption is particularly challenging because predictor–predictand relationships may not remain stationary under climate change \citep{dixon2013examining}. Establishing the trust necessary to apply emulators beyond their original development context requires an explicit assessment of their ability to generalize across temporal and spatial scales, scenarios, climate models, variables, or other aspects \citep{gutmann2014intercomparison, rampal_extrapolation_2024, tebaldi2025emulators}. 
\subsection{Accessibility}
Even when emulators have demonstrated skill and gained trust, practical barriers often hinder their adoption. A lack of standardized model access, reproducible workflows, and interoperable software can prevent ML tools from being integrated into established research pipelines
\footnote{One way to lower this barrier would be to integrate ML-generated data into established community infrastructures like the Earth System Grid Federation (ESCF), allowing emulators to be trained on standardized datasets and their outputs to be distributed in consistent, interoperable formats.}.
Therefore, shifting toward more open release practices is a prerequisite for moving emulators from individual experiments to widely adopted tools. 
In this context, documentation plays an important role. 
To ensure accessibility and support meaningful co-evaluation, developers should clearly document model assumptions, limitations, and intended use cases, avoiding unnecessary technical terms. We recommend adopting interoperable data formats, such as netCDF or zarr, to ensure emulator outputs can be seamlessly integrated into established research workflows. 
Furthermore, leveraging standardized frameworks like Anemoi\footnote{Anemoi provides a open-source ecosystem for data preprocessing, model training and inference, with the framework currently including several AI weather forecast models. Establishing similar unified frameworks for other fields such as GCM or RCM model emulation would be a major step forward.} \citep{lang2024aifsecmwfsdatadriven} can significantly simplify the adoption of these methods. 

While unconventional, we believe it would be beneficial for ML researchers to explicitly state their interest in potential use cases within their publications. Such statements could foster impactful collaborations, ensuring that model development yields results that are directly meaningful to the applied community. 

\section{Successfully applying an emulator}
While \Cref{sec:development} outlines the responsibilities of emulator developers, successful collaboration depends equally on the engagement of applied scientists.
But at which point exactly should researchers from the two fields interact?
\Cref{fig:combinedworkflow} suggests a simple solution: instead of using observational records, reanalysis products, or model output, applied researchers could use emulator-generated data. However, this rarely happens in practice \citep[\eg][]{fowler2025downscaling}. 
The issue is usually not the quality of emulator-generated data, as recent ML emulators already show strong performance \citep{van2023deep, kendon2025potential, schillinger2025enscale, addison_machine_2026, legasa_regional_2026}. 

In theory, the application of emulators is relatively straightforward. Since emulators function essentially as a new data source, their validation and usage should be performed in a manner consistent with traditional physical data. Selecting a specific emulator is highly dependent on the research question; therefore, we refrain from offering universal recommendations here. Instead, we argue that the final selection should be guided by the three principles established earlier: adequacy for purpose, trustworthiness, and accessibility.
In practice, these translate into three practical questions for the applied researcher: Can this emulator provide the specific data needed to answer my question? Do I trust its performance? Do I have the technical capabilities to implement the model or download the resulting data? In addition to this, there is a psychological barrier: researchers must have the confidence to move beyond traditional physical models and decide how an emulator should be positioned alongside them. Furthermore, the emulator must demonstrate a tangible advantage over established methods. While this is partially addressed by the identified added value, the decision ultimately hinges on a cost-benefit analysis, and the perceived benefits must clearly outweigh the effort required to implement the new approach.

Ultimately, emulator selection is a process of matching the specific requirements of a research question with the proven properties of a model. A researcher interested in quantifying the uncertainty of extremes, for example, will only deem an emulator useful and trustworthy if its distributional properties have been rigorously tested. While we recommend that ML researchers perform extensive, purpose-oriented validation, we acknowledge that it is impossible to address every potential research question when presenting a new method. However, if there is no evidence that a model is likely to fit the specific scope, the perceived risk of wasting the applied researcher's time may outweigh the potential benefits.  

After deciding on an emulator, the applied researcher should further validate that it adequately captures the phenomena they are interested in. Much like the standard procedure for selecting physical climate model data, this process usually involves validation against observations or reanalysis data. After this validation step, the data can be used in the same way as validated output from a physical climate model.

\subsection{What are limitations and future possibilities?}
The field of climate emulation is moving fast, with initiatives like AIMIP Phase 1 \citep{henn2026aimip} and CORDEX-ML-Bench \citep{rampal_cordexbench_2026} being developed to standardize ML models just as CMIP did for physical climate science. Ultimately, the impact of these tools will depend on whether we can actually bring the communities together. Progress will likely be driven by a new generation of researchers who are fluent in both domains; supported by shared PhD student supervision and interdisciplinary grants. While some researchers already define themselves as working at this intersection (including the authors of this paper), this group will likely grow and diversify. As the field matures, these researchers will move into more nuanced niches, refining the specialized methodologies required to bridge the gap between machine learning and climate science. By establishing a framework where applied researchers define the critical (data) challenges and methodological experts develop the solutions, the community can move towards impactful future research.

New methods, like ML emulators, often become attractive whenever traditional models reach their limits. However, this potential should not lead to unrealistic expectations. 
Some applications are currently difficult to address reliably with ML-based climate model emulators. Counterfactual analyses, for example, are more naturally implemented in physical simulators, where model components or forcings can be systematically modified. A related challenge arises when emulators are used to \textit{project} future climate states. Unlike physical climate models, climate model emulators do not explicitly represent Earth system feedbacks, such as sea-ice-albedo interactions, and are therefore unable to sufficiently generalize under changing climatic conditions \citep{ullrich2025recommendations}. This differs from weather emulation, where short-term forecasts can often be made from the current atmosphere \citep[\eg][]{price2025probabilistic}. 

Similar caution is needed in the context of \textit{climate data poverty}. In regions or applications where reliable observations are sparse, synthetic high-resolution data generated by ML emulators may appear to offer a solution. Yet, if these emulators are trained on biased or incomplete climate model output, reanalysis products, or observations, they may reproduce or amplify existing biases rather than resolve them. In doing so, they risk projecting current data limitations onto already under-served regions, thereby widening knowledge gaps and reinforcing climate data poverty. This is particularly concerning in parts of the Global South, where the lack of high-quality observational data already poses a major challenge for climate research \citep{otto2020challenges}, despite high vulnerability to climate change \citep{campbell2007climate}.

At the same time, these current limitations also point to where the potential of ML emulators may be greatest. If developed carefully, they could help address geographical inequities in climate information by supporting regions where observations are sparse, computational resources are limited, or traditional modeling approaches are difficult to apply. Recent work in the ML community is beginning to address such settings explicitly \citep[\eg][]{harder2026benchmarking}.

\section{Conclusion}
Since emulator development and application are often conducted by separate research communities, the success of these tools depends on bridging the gap between them. We argue that when developers keep specific applications in mind from the outset, the transition to practical use becomes significantly more seamless. To support this, developers should prioritize adequacy for a scientific purpose, trustworthiness, and accessibility of their models, as these are the core values that applied researchers require to trust and apply a model. Ultimately, seeing clear added value and feeling confident in these principles, may motivate applied researchers to move beyond traditional approaches. This hopefully allows for the creation of a connection that transforms a theoretical model into a standard scientific tool.



\paragraph{Funding Statement}
N.E., M.Sa., and M.Sc. are part of SPEED2ZERO, a Joint Initiative co-financed by the ETH Board.
L.S. was supported by the \textit{Deutsche Forschungsgemeinschaft} (DFG, German Research Foundation) under Germany's Excellence Strategy – EXC number 2064/1 – Project number 390727645. The authors thank the International Max Planck Research School for Intelligent Systems (IMPRS-IS) for supporting L.S.. M.L. received funding managed by the \textit{Agence Nationale de la Recherche} under France 2030 bearing the references ANR-22-EXTR-0005 (TRACCS-PC4-EXTENDING project) and ANR-22-EXTR-0011 (TRACCS-PC10-LOCALISING project). P.L.B receives funding from ARIA and DSIT and Pillar VC under the Encode: Al for Science Fellowship. R.B. is funded by the Deutsche Forschungsgemeinschaft (DFG, German Research Foundation) – Project number 537063406.



\paragraph{Author Contributions}
Conceptualization: L.S.*, N.E.*, V.B., P.B., R.B., M.L., M.Sa., M.Sc.. Methodology: L.S.*; N.E*., V.B.,P.B., R.B., M.L., M.Sa., M.Sc.. Writing - original draft: L.S.; N.E. Writing - review \& editing: V.B.,P.B., R.B., M.L., M.Sa., M.Sc..  \\
* Denotes a lead role; all other listed authors contributed in a supporting role and are ordered alphabetically.
\paragraph{Acknowledgements} 
We thank Anastase Charantonis, Julie Keisler, Frieder Loer, and Renu Singh for their valuable input and discussions.
\bibliographystyle{plainnat}
\bibliography{example}
\newpage
\begin{appendices}
\section{Supplementary Materials} \label{appendix:A}
\setcounter{figure}{0}
\counterwithin{figure}{section}
\subsection{Analysis of ACE2 References}
\label{sec:ace2_references}
To analyze of how prior literature builds upon the ACE2 emulator framework \citep{watt2025ace2}, we conducted a systematic review of the 60 papers (as of 15 June 2026) that reference the original ACE2 publication. The main goal was to detect papers that use ACE2 within a practical application, \ie to generate new knowledge beyond that of existing data. 

As summarized in \Cref{tab:ace2_analysis}, the vast majority of publications (33 papers) cite ACE2 contextually without applying it directly. Furthermore, eleven papers use ACE2 for benchmarking purposes without introducing novel insights that extend beyond established physical baselines. However, we want to highlight that these papers can often serve as an in-between step before generating new knowledge. For instance, see the example of tropical cyclogenesis described in the table. Five papers are literature reviews, and eight papers reference ACE2 while introducing an entirely new model (some of which are based on ACE2). Only two papers ($3.3\%$) of the analyzed literature transition ACE2 into a concrete, practical application that generates novel insights extending beyond established physical baselines.

\begin{table}[htbp]
\centering
\caption{Classification and Qualitative Analysis of Papers Referencing ACE2.}
\label{tab:ace2_analysis}
\begin{tabular}{lcp{10cm}}
\hline
\textbf{Category} & \textbf{Count} & \textbf{Description / Core Engagement} \\ \hline
Citation Only & 33 & Contextual reference only; does not use or implement the ACE2 framework. \\
Benchmark & 12 & Uses ACE2 as a baseline for performance comparison without generating insights beyond the physical baseline. \\
Model development & 8 & Introduces a new model or architecture  referencing ACE2. It might be noted that three of these papers \citep{clark2025ace2, perkins2026hiroacefastskillfulai,ace2Samudr} were published in part by members of the initial ACE2 team.\\
Review & 5 & Review or perspective papers, synthesizing existing literature. \\
Practical Application & 2 & Actively implements the ACE2 framework to solve domain-specific applications.
\begin{itemize}
    \item Tropical cyclogenesis:  more robust relationship between tropical cyclogenesis and Kelvin waves by analyzing ACE2-ERA5 instead of observations \citep{chien2026modulation}. A previous paper by the authors demonstrated that the relationship between tropical cyclogenesis and the large-scale environment is reasonably represented in ACE2-ERA5 \citep{chien2025modulation}.  
    \item Very extreme precipitation and temperature quantification: huge ensembles can be practically used alongside threshold-exceedance extreme value techniques to reliably estimate rare, high-consequence precipitation and temperature extremes \cite{paciorek2026quantifying}.
\end{itemize}
\\ \hline
Total & \textbf{60} & \\ \hline
\end{tabular}
\end{table}
\begin{figure}[h!]
    \centering
\begin{tcolorbox}[
    colback=gray!7, 
    colframe=gray!70, 
    width=\textwidth, 
    arc=2mm, 
    boxrule=1pt,
    title=\textbf{Checklist for Emulator Development},
    coltitle=white,
    colbacktitle=gray!60,
    fonttitle=\sffamily\bfseries,
    left=2mm,
    right=2mm,
    top=2mm,
    bottom=2mm,
    toptitle=3mm,
    bottomtitle=3mm
]
    \begin{enumerate}[label=\textbf{\arabic*.}, leftmargin=1.5em, itemsep=0.5em]
        \item \textbf{Formulate a scientifically motivated task.}
        \begin{itemize}[label=$\square$, nosep]

            \item Define the learning objective based on a scientifically motivated problem.
            \begin{itemize}
                \item Describe the downstream application or research context.
                \item Identify which aspects are relevant for the application, e.g. variability, extremes, temporal dynamics.
                \item Derive the modeling objective from this motivation.
            \end{itemize}
            \item Justify the use of an emulator for this problem.
            \begin{itemize}
                \item Explain why an emulator is needed in this context.
                \item Clarify the expected added value relative to the input.
                \item Explain which signal(s) is expected to drive the outputs and why it matters for the application.
            \end{itemize}
            \item Choose datasets consistent with the scientific problem.

            \begin{itemize}
                \item Ensure that dataset pairing matches the scientific problem.
                \item Align resolution, spatial and temporal domain with the relevant physical processes and scientific problem.
                \item Verify that predictor variables are physically meaningful.
                \item Check that training data covers the relevant regimes, e.g. scenarios, seasons, climate states.
                \item Document known dataset biases and limitations and consider their implications.
                \item Document all pre-processing steps transparently and reproducibly.
            \end{itemize}
        
        \end{itemize}
        \item \textbf{Evaluate emulator for its intended purpose.}
        \begin{itemize}[label=$\square$, nosep]
            \item Select and justify metrics in terms of the application.
            \item Demonstrate that the emulator credibly reproduces reference data, including climatology and internal variability.
            \item Evaluate robustness and generalization across dimensions relevant to use.
        \end{itemize}

        \item \textbf{Demonstrate context-dependent added value.} 
        \begin{itemize}[label=$\square$, nosep]
            \item Define what added value means for the application.
            \item Measure added value relative to a physically meaningful reference.
            \item Use metrics that matter for the application.
        \end{itemize}

        \item \textbf{Discuss modeling assumptions and uncertainties.}
        \begin{itemize}[label=$\square$, nosep]
            \item Discuss modeling assumptions explicitly.
            \item Specify which sources of uncertainty the emulator represents and which not.
            \item Address known failure modes and robustness risks.
            \item State implications for transferability: under which conditions should the emulator not be trusted? 
        \end{itemize}

            \item \textbf{Ensure transparency, reproducibility, and accessibility}
        \begin{itemize}[label=$\square$, nosep]
            \item Provide open access to the full workflow (preprocessing pipeline, model architecture, trained weights, input/output data)
            \item Use interoperable formats.
            \item Clearly document intended use, scope, and known limitations in a way that is accessible also to non-ML users.
        \end{itemize}

    \end{enumerate}
    \end{tcolorbox}
    \caption{Checklist for emulator development.}
    \label{fig:checklist}
\end{figure}
\end{appendices}
\end{document}